\documentclass[11pt,a4paper]{article}
\usepackage[hyperref]{emnlp2020}
\usepackage{times}
\usepackage{latexsym}

\usepackage{graphicx}
\usepackage{subfigure}
\usepackage{float}
\usepackage{amsmath}
\usepackage{amssymb}
\usepackage{multirow}
\usepackage{booktabs}
\usepackage{url}
\usepackage{array}
\usepackage{enumitem}
\usepackage{algorithm}
\usepackage{algorithmic}
\usepackage{pifont}
\usepackage{CJKutf8}

\usepackage{microtype}

\aclfinalcopy 

\title{Revisiting Pre-trained Models for Chinese Natural Language Processing}

\author{Yiming Cui$^{1,2}$, Wanxiang Che$^1$, Ting Liu$^1$, Bing Qin$^1$, Shijin Wang$^{2,3}$, Guoping Hu$^2$\\
{$^1$Research Center for Social Computing and Information Retrieval (SCIR),}\\
{Harbin Institute of Technology, Harbin, China}\\
{$^2$State Key Laboratory of Cognitive Intelligence, iFLYTEK Research, China}\\
{$^3$iFLYTEK AI Research (Hebei), Langfang, China} \\
{$^1$\tt \{ymcui,car,tliu,qinb\}@ir.hit.edu.cn}\\
{$^{2,3}$\tt\{ymcui,sjwang3,gphu\}@iflytek.com}\\  
}

\date{}

\begin{document}
\begin{CJK*}{UTF8}{gbsn}
\maketitle

\begin{abstract}
Bidirectional Encoder Representations from Transformers (BERT) has shown marvelous improvements across various NLP tasks, and consecutive variants have been proposed to further improve the performance of the pre-trained language models.
In this paper, we target on revisiting Chinese pre-trained language models to examine their effectiveness in a non-English language and release the Chinese pre-trained language model series to the community.
We also propose a simple but effective model called MacBERT, which improves upon RoBERTa in several ways, especially the masking strategy that adopts {\em MLM as correction} (Mac).
We carried out extensive experiments on eight Chinese NLP tasks to revisit the existing pre-trained language models as well as the proposed MacBERT.
Experimental results show that MacBERT could achieve state-of-the-art performances on many NLP tasks, and we also ablate details with several findings that may help future research.\footnote{\url{https://github.com/ymcui/MacBERT}}
\end{abstract}

\section{Introduction}
Bidirectional Encoder Representations from Transformers (BERT) \citep{devlin-etal-2019-bert} has become enormously popular and has proven to be effective in recent natural language processing studies, which utilizes large-scale unlabeled training data and generates enriched contextual representations.
As we traverse several popular machine reading comprehension benchmarks, such as SQuAD \citep{rajpurkar-etal-2018-know}, CoQA \citep{reddy2019coqa}, QuAC \citep{choi-etal-2018-quac}, NaturalQuestions \citep{kwiatkowski2019natural}, RACE \citep{lai-etal-2017}, we can see that most of the top-performing models are based on BERT and its variants \citep{dai-etal-2019-transformer,zhang2019dual,ran2019option}, demonstrating that the pre-trained language models have become new fundamental components in natural language processing field.

Starting from BERT, the community have made great and rapid progress on optimizing the pre-trained language models, such as ERNIE \citep{sun2019ernie}, XLNet \citep{yang2019xlnet}, RoBERTa \citep{liu2019roberta}, SpanBERT \citep{joshi2019spanbert}, ALBERT \citep{lan2019albert}, ELECTRA \citep{clark2020electra}, etc. 
However, training Transformer-based \citep{vaswani2017attention} pre-trained language models are not as easy as we used to train word embeddings or other traditional neural networks. 
Typically, training a powerful BERT-large model, which has 24-layer Transformer with 330 million parameters, to convergence needs high-memory computing devices, such as TPU, which is very expensive. 
On the other hand, though various pre-trained language models have been released, most of them are based on English, and there are few efforts on building powerful pre-trained language models on other languages. 

\begin{table*}[htbp]
\small
\begin{center}
\begin{tabular}{l c c c c c c | c}
\toprule
& \bf BERT & \bf ERNIE & \bf XLNet & \bf RoBERTa & \bf ALBERT & \bf ELECTRA & \bf MacBERT \\
\midrule
Type & AE & AE & AR & AE & AE & AE & AE  \\
Embeddings & T/S/P & T/S/P & T/S/P & T/S/P & T/S/P & T/S/P & T/S/P \\
Masking & T & T/E/Ph & - & T & T & T & WWM/NM \\
LM Task & MLM & MLM & PLM & MLM & MLM & Gen-Dis & Mac \\
Paired Task & NSP & NSP & - & - & SOP & - & SOP \\
\bottomrule
\end{tabular}
\end{center}
\caption{\label{model-comparison} Comparisons of the pre-trained language models. (AE: Auto-Encoder, AR: Auto-Regressive, T: Token, S: Segment, P: Position, W: Word, E: Entity, Ph: Phrase, WWM: Whole Word Masking, NM: N-gram Masking, NSP: Next Sentence Prediction, SOP: Sentence Order Prediction, MLM: Masked LM, PLM: Permutation LM, Mac: MLM as correction)}
\end{table*}

In this paper, we aim to build Chinese pre-trained language model series and release them to the public for facilitating the research community, as Chinese and English are among the most spoken languages in the world.
We revisit the existing popular pre-trained language models and adjust them to the Chinese language to see if these models generalize well in the language other than English.
Besides, we also propose a new pre-trained language model called MacBERT, which replaces the original MLM task into {\bf M}LM {\bf a}s {\bf c}orrection (Mac) task and mitigates the discrepancy of the pre-training and fine-tuning stage.
Extensive experiments are conducted on eight popular Chinese NLP datasets, ranging from sentence-level to document-level, such as machine reading comprehension, text classification, etc.
The results show that the proposed MacBERT could give significant gains in most of the tasks against other pre-trained language models, and detailed ablations are given to better examine the composition of the improvements.
The contributions of this paper are listed as follows.

\begin{itemize}
	\item Extensive empirical studies are carried out to revisit the performance of Chinese pre-trained language models on various tasks with careful analyses.
	\item We propose a new pre-trained language model called MacBERT that mitigates the gap between the pre-training and fine-tuning stage by masking the word with its similar word, which has proven to be effective on down-stream tasks.
	\item To further accelerate future research on Chinese NLP, we create and release the Chinese pre-trained language model series to the community.
\end{itemize}

\section{Related Work}
In this section, we revisit the techniques of the representative pre-trained language models in the recent natural language processing field.
The overall comparisons of these models, as well as the proposed MacBERT, are depicted in Table \ref{model-comparison}.
We elaborate on their key components in the following subsections.

\subsection{BERT}
BERT (Bidirectional Encoder Representations from Transformers) \citep{devlin-etal-2019-bert} has proven to be successful in natural language processing studies. 
BERT is designed to pre-train deep bidirectional representations by jointly conditioning on both left and right context in all Transformer layers.
Primarily, BERT consists of two pre-training tasks: Masked Language Model (MLM) and Next Sentence Prediction (NSP).
\begin{itemize}
	\item {\bf MLM}: Randomly masks some of the tokens from the input, and the objective is to predict the original word based only on its context.
	\item {\bf NSP}: To predict whether sentence {\em B} is the next sentence of {\em A}.
\end{itemize}

Later, they further proposed a technique called whole word masking (wwm) for optimizing the original masking in the MLM task.
In this setting, instead of randomly selecting WordPiece \citep{wu2016google} tokens to mask, we always mask all of the tokens corresponding to a whole word at once. 
This will explicitly force the model to recover the whole word in the MLM pre-training task instead of just recovering WordPiece tokens \citep{chinese-bert-wwm}, which is much more challenging.
As the whole word masking only affects the masking strategy of the pre-training process, it would not bring additional burdens on down-stream tasks.
Moreover, as training pre-trained language models are computationally expensive, they also release all the pre-trained models as well as the source codes, which stimulates the community to have great interests in the research of pre-trained language models.

\subsection{ERNIE}
ERNIE (Enhanced Representation through kNowledge IntEgration) \citep{sun2019ernie} is designed to optimize the masking process of BERT, which includes entity-level masking and phrase-level masking. 
Different from selecting random words in the input, entity-level masking will mask the named entities, which are often formed by several words.
Phrase-level masking is to mask consecutive words, which is similar to the N-gram masking strategy \citep{devlin-etal-2019-bert,joshi2019spanbert}.\footnote{Though N-gram masking was not included in \citet{devlin-etal-2019-bert}, according to their model name in the SQuAD leaderboard, we often admit their credit towards this method.}. 

\subsection{XLNet}
\citet{yang2019xlnet} argues that the existing pre-trained language models that are based on autoencoding, such as BERT, suffer from the discrepancy of the pre-training and fine-tuning stage because the masking symbol {\tt [MASK]} will never appear in the fine-tuning stage.
To alleviate this problem, they proposed XLNet, which was based on Transformer-XL \citep{dai-etal-2019-transformer}. 
XLNet mainly modifies in two ways. 
The first is to maximize the expected likelihood over all permutations of the factorization order of the input, where they called the Permutation Language Model (PLM). 
Another is to change the autoencoding language model into an autoregressive one, which is similar to the traditional statistical language models.\footnote{We also trained Chinese XLNet, but it only shows competitive performance on reading comprehension datasets. We've included these results in the Appendix.}

\begin{figure*}[tbp]
\centering\tiny
        \begin{tabular}{l l l}
        \toprule
        & {\bf Chinese} & {\bf English} \\
        \midrule
	{\bf Original Sentence} & 使用语言模型来预测下一个词的概率。 & we use a language model to predict the probability of the next word.\\
        {\bf + CWS} & 使用~语言~{\bf 模型}~来~{\bf 预测}~下~一个~词~的~{\bf 概率}~。 & - \\
        {\bf + BERT Tokenizer} & 使~用~语~言~{\bf 模~型}~来~{\bf 预~测}~下~一~个~词~的~{\bf 概~率}~。 & we~use~a~language~{\bf model}~to~{\bf pre~\#\#di~\#\#ct}~the~{\bf pro~\#\#ba~\#\#bility}~of~the~next~word~.  \\
        \midrule
        {\bf Original Masking} & 使~用~语~言~{\bf [M]~型}~来~{\bf [M]~测}~下~一~个~词~的~{\bf 概~率}~。 & we~use~a~language~{\bf [M]}~to~{\bf [M]~\#\#di~\#\#ct}~the~{\bf pro~[M]~\#\#bility}~of~the~next~word~.  \\
        {\bf + WWM} & 使~用~语~言~{\bf [M]~[M]}~来~{\bf [M]~[M]}~下~一~个~词~的~{\bf 概~率}~。 & we~use~a~language~{\bf [M]}~to~{\bf [M]~[M]~[M]}~the~{\bf [M]~[M]~[M]}~of~the~next~word~.  \\
        {\bf ++ N-gram Masking} & 使~用~{\bf [M]~[M]~[M]~[M]}~来~{\bf [M]~[M]}~下~一~个~词~的~{\bf 概~率}~。  & we~use~a~{\bf [M]}~{\bf [M]}~to~{\bf [M]~[M]~[M]}~the~{\bf [M]~[M]~[M]}~{\bf [M]~[M]}~next~word~.  \\
        {\bf +++ Mac Masking} & 使~用~{\bf 语~法~建~模}~来~{\bf 预~见}~下~一~个~词~的~{\bf 几~率}~。  & we~use~a~{\bf text}~{\bf system}~to~{\bf ca~\#\#lc~\#\#ulate}~the~{\bf po~\#\#si~\#\#bility}~of~the~next~word~.  \\
        \bottomrule
        \end{tabular}
\caption{\label{example} Examples of different masking strategies. }
\end{figure*}

\subsection{RoBERTa}
RoBERTa (Robustly Optimized BERT Pretraining Approach) \citep{liu2019roberta} aims to adopt original BERT architecture but make much more precise modifications to show the powerfulness of BERT, which was underestimated. 
They carried out careful comparisons of various components in BERT, including the masking strategies, training steps, etc. 
After thorough evaluations, they came up with several useful conclusions to make BERT more powerful, mainly including 1) training longer with bigger batches and longer sequences over more data; 2) removing the next sentence prediction and using dynamic masking.

\subsection{ALBERT}
ALBERT (A Lite BERT) \citep{lan2019albert} primarily tackles the problems of higher memory assumption and slow training speed of BERT.
ALBERT introduces two parameter reduction techniques.
The first one is the factorized embedding parameterization that decomposes the embedding matrix into two small matrices.
The second one is the cross-layer parameter sharing that the Transformer weights are shared across each layer of ALBERT, which will significantly reduce the parameters.
Besides, they also proposed the sentence-order prediction (SOP) task to replace the traditional NSP pre-training task.

\subsection{ELECTRA}
ELECTRA (Efficiently Learning an Encoder that Classifiers Token Replacements Accurately) \citep{clark2020electra} employs a new generator-discriminator framework that is similar to GAN \citep{goodfellow-gan-nips2014}.
The generator is typically a small MLM that learns to predict the original words of the masked tokens.
The discriminator is trained to discriminate whether the input token is replaced by the generator.
Note that, to achieve efficient training, the discriminator is only required to predict a binary label to indicate ``replacement'', unlike the way of MLM that should predict the exact masked word.
In the fine-tuning stage, only the discriminator is used.

\section{Chinese Pre-trained Language Models}
While we believe most of the conclusions in the previous works are true in English condition, we wonder if these techniques still generalize well in other languages.
In this section, we illustrate how the existing pre-trained language models are adapted for the Chinese language.
Furthermore, we also propose a new model called MacBERT, which adopts the advantages of previous models as well as newly designed components.
Note that, as these models are all originated from BERT without changing the nature of the input, no modification should be made to adapt to these models in the fine-tuning stage, which is very flexible for replacing one another.

\subsection{BERT-wwm \& RoBERTa-wwm}
In the original BERT, a WordPiece tokenizer \citep{wu2016google} was used to split the text into WordPiece tokens, where some words will be split into several small fragments.
The whole word masking (wwm) mitigate the drawback of masking only a part of the whole word, which is easier for the model to predict.
In Chinese condition, WordPiece tokenizer no longer split the word into small fragments, as Chinese characters are not formed by alphabet-like symbols.
We use the traditional Chinese Word Segmentation (CWS) tool to split the text into several words.
In this way, we could adopt whole word masking in Chinese to mask the word instead of individual Chinese characters.
For implementation, we strictly followed the original whole word masking codes and did not change other components, such as the percentage of word masking, etc.
We use LTP \citep{che2010ltp} for Chinese word segmentation to identify the word boundaries.
Note that the whole word masking only affects the selection of the masking tokens in the pre-training stage. 
The input of BERT still uses WordPiece tokenizer to split the text, which is identical to the original BERT.

Similarly, whole word masking could also be applied on RoBERTa, where the NSP task is not adopted.
An example of the whole word masking is depicted in Figure \ref{example}.

\begin{table*}[htbp]
\small
\begin{center}
\begin{tabular}{c l c | c c c c | c c c}
\toprule
\bf Task & \bf Dataset & \bf Domain & \bf MaxLen & \bf Batch & \bf Epoch & \bf InitLR & \bf Train \# & \bf Dev \# & \bf Test \#  \\
\midrule
\multirow{3}*{\bf MRC} & CMRC 2018  & Wikipedia	 & 512 & 64 & 2 & 3e-5 & 10K & 3.2K & 4.9K  \\
& DRCD 		 & Wikipedia & 512 & 64 & 2 & 3e-5 & 27K & 3.5K & 3.5K  \\
& CJRC 			 & Law  & 512 & 64 & 2 & 4e-5 & 10K & 3.2K & 3.2K  \\
\midrule
\multirow{2}*{\bf SSC} & ChnSentiCorp 	 & Various & 256 & 64 & 3 & 2e-5 & 9.6K & 1.2K & 1.2K \\
 & THUCNews 		 & News & 512 & 64 & 3 & 2e-5 & 50K & 5K & 10K  \\
\midrule
\multirow{3}*{\bf SPC} & XNLI	& Various & 128 & 64 & 2 & 3e-5 & 392K & 2.5K & 5K  \\
& LCQMC 			 & Zhidao & 128 & 64 & 3 & 2e-5 & 240K & 8.8K & 12.5K  \\
& BQ Corpus	  	 	& QA & 128 & 64 & 3 & 3e-5 & 100K & 10K & 10K  \\
\bottomrule
\end{tabular}
\end{center}
\caption{\label{hyper} Data statistics and hyper-parameter settings for different fine-tuning tasks.}
\end{table*}

\subsection{MacBERT}
In this paper, we take advantage of previous models and propose a simple modification that leads to significant improvements on fine-tuning tasks, where we call this model as {\bf MacBERT} ({\bf M}LM {\bf a}s {\bf c}orrection {\bf BERT}).
MacBERT shares the same pre-training tasks as BERT with several modifications.
For the MLM task, we perform the following modifications.
\begin{itemize}
	\item We use whole word masking as well as N-gram masking strategies for selecting candidate tokens for masking, with a percentage of 40\%, 30\%, 20\%, 10\% for word-level unigram to 4-gram.
	\item Instead of masking with {\tt [MASK]} token, which never appears in the fine-tuning stage, we propose to use similar words for the masking purpose. A similar word is obtained by using {\em Synonyms} toolkit \citep{Synonyms:hain2017}, which is based on word2vec \citep{mikolov-etal-2013} similarity calculations. If an N-gram is selected to mask, we will find similar words individually. In rare cases, when there is no similar word, we will degrade to use random word replacement.
	\item We use a percentage of 15\% input words for masking, where 80\% will replace with similar words, 10\% replace with a random word, and keep with original words for the rest of 10\%.
\end{itemize}
For the NSP-like task, we perform sentence-order prediction (SOP) task as introduced by ALBERT \citep{lan2019albert}, where the negative samples are created by switching the original order of two consecutive sentences.
We ablate these modifications in Section \ref{effect-macbert} to better demonstrate the contributions of each component.

\begin{table*}[htbp]
\small
\begin{center}
\begin{tabular}{l c c c c c }
\toprule
 	& \bf BERT & \bf BERT-wwm & \bf RoBERTa-wwm & \bf ELECTRA & \bf MacBERT \\
\midrule
Word \#			& 0.4B & 5.4B  & 5.4B & 5.4B & 5.4B \\
Vocab \#			& 21,128 & 21,128 & 21,128 & 21,128 & 21,128 \\
Hidden Activation	& GeLU & GeLU  & GeLU & GeLU & GeLU  \\
Optimizer 			& AdamW & LAMB & AdamW & AdamW & LAMB \\
Training Steps		& ? & 2M & 1M & 2M & 1M \\
Init Checkpoint 	& random & BERT & BERT & random & BERT\\
\bottomrule
\end{tabular}
\end{center}
\caption{\label{comparison-base} Training details of Chinese pre-trained language models.}
\end{table*}

\section{Experimental Setups}
\subsection{Setups for Pre-Trained Language Models}
We downloaded Wikipedia dump\footnote{https://dumps.wikimedia.org/zhwiki/latest/} (as of March 25, 2019), and pre-processed with {\tt WikiExtractor.py} as suggested by \citet{devlin-etal-2019-bert}, resulting in 1,307 extracted files. 
We use both Simplified and Traditional Chinese in this dump.
After cleaning the raw text (such as removing {\tt html} tagger) and separating the document, we obtain about 0.4B words.
As Chinese Wikipedia data is relatively small, besides Chinese Wikipedia, we also use extended training data for training these pre-trained language models (mark with {\tt ext} in the model name).
The in-house collected extended data contains encyclopedia, news, and question answering web, which has 5.4B words and is over ten times bigger than the Chinese Wikipedia.
Note that we always use extended data for MacBERT, and omit the {\tt ext} mark.
In order to identify the boundary of Chinese words, we use LTP \citep{che2010ltp} for Chinese word segmentation.
We use official {\tt create\_pretraining\_data.py} to convert raw input text to the pre-training examples.

To better acquire the knowledge from the existing pre-trained language model, we did NOT train our base-level model from scratch but the official Chinese BERT-base, inheriting its vocabulary and weight.
However, for the large-level model, we have to train from scratch but still using the same vocabulary provided by the base-level model.

For training BERT series, we adopt the scheme of training on a maximum length of 128 tokens then on 512, suggested by \citet{devlin-etal-2019-bert}.
However, we empirically found that this will result in insufficient adaptation for the long-sequence tasks, such as reading comprehension.
In this context, for RoBERTa and MacBERT, we directly use a maximum length of 512 throughout the pre-training process, which was adopted in \citet{liu2019roberta}.
For the batch size less than 1024, we adopt the original \textsc{Adam} \citep{kingma2014adam} with weight decay optimizer in BERT for optimization, and use \textsc{LAMB} optimizer \citep{you2019reducing} for better scalability in larger batch.
The pre-training was either done on a single Google Cloud TPU\footnote{https://cloud.google.com/tpu/} v3-8 (equals to a single TPU) or TPU Pod v3-32 (equals to 4 TPUs), depending on the magnitude of the model. 
Specifically, for MacBERT-large, we trained for 2M steps with a batch size of 512 and an initial learning rate of 1e-4.

The training details are shown in Table \ref{comparison-base}. 
For clarity, we do not list `{\tt ext}' models, where the other parameters are the same with the one that is not trained on extended data.

\begin{table*}[htbp]
\small
\begin{center}
\begin{tabular}{p{3.5cm} c c c c c c}
\toprule
\multirow{2}*{\bf CMRC 2018} & \multicolumn{2}{c}{\centering \bf Dev} & \multicolumn{2}{c}{\centering \bf Test} & \multicolumn{2}{c}{\centering \bf Challenge} \\
& \bf EM & \bf F1 & \bf EM & \bf F1  & \bf EM & \bf F1  \\
\midrule
BERT   			& 65.5 \tiny(64.4) & 84.5 \tiny(84.0) & 70.0 \tiny(68.7) & 87.0 \tiny(86.3)  & 18.6 \tiny(17.0) & 43.3 \tiny(41.3) \\
BERT-wwm     	& 66.3 \tiny(65.0) & 85.6 \tiny(84.7) & 70.5 \tiny(69.1) & 87.4 \tiny(86.7)  & 21.0 \tiny(19.3) & 47.0 \tiny(43.9) \\
BERT-wwm-ext	& 67.1 \tiny(65.6) & 85.7 \tiny(85.0) & 71.4 \tiny(70.0) & 87.7 \tiny(87.0) & 24.0 \tiny(20.0) & 47.3 \tiny(44.6) \\
RoBERTa-wwm-ext & 67.4 \tiny(66.5) & 87.2 \tiny(86.5) & 72.6 \tiny(71.4) & 89.4 \tiny(88.8) & 26.2 \tiny(24.6) & 51.0 \tiny(49.1) \\
ELECTRA-base & 68.4 \bf\tiny(68.0) & 84.8 \tiny(84.6) & 73.1 \bf\tiny(72.7) & 87.1 \tiny(86.9) & 22.6 \tiny(21.7) & 45.0 \tiny(43.8) \\
\bf MacBERT-base & {\bf 68.5} \tiny(67.3) & \bf 87.9 \tiny(87.1) & {\bf 73.2} \tiny(72.4) & \bf 89.5 \tiny(89.2) & \bf 30.2 \tiny(26.4) & \bf 54.0 \tiny(52.2) \\
\midrule
ELECTRA-large & 69.1 \tiny(68.2) & 85.2 \tiny(84.5) & 73.9 \tiny(72.8) & 87.1 \tiny(86.6) & 23.0 \tiny(21.6) & 44.2 \tiny(43.2) \\
RoBERTa-wwm-ext-large & 68.5 \tiny(67.6) & 88.4 \tiny(87.9) & 74.2 \tiny(72.4) & 90.6 \tiny(90.0) & 31.5 \bf \tiny(30.1) & 60.1 \tiny(57.5)  \\
\bf MacBERT-large 		& \bf 70.7 \tiny(68.6) & \bf 88.9 \tiny(88.2) & \bf 74.8 \tiny(73.2) & \bf 90.7 \tiny(90.1) & {\bf 31.9} \tiny(29.6) & \bf 60.2 \tiny(57.6)  \\
\bottomrule
\end{tabular}
\end{center}
\caption{\label{result-cmrc2018} Results on CMRC 2018 (Simplified Chinese). The average scores of 10 independent runs are depicted in brackets. Overall best performances are depicted in boldface (base-level and large-level are marked individually).}
\end{table*}

\subsection{Setups for Fine-tuning Tasks}
To thoroughly test these pre-trained language models, we carried out extensive experiments on various natural language processing tasks, covering a wide spectrum of text length, i.e., from sentence-level to document-level. Task details are shown in Table \ref{hyper}.
Specifically, we choose the following eight popular Chinese datasets.
\begin{itemize}[leftmargin=*]
	\item {\bf Machine Reading Comprehension (MRC)}: CMRC 2018 \citep{cui-emnlp2019-cmrc2018}, DRCD \citep{shao2018drcd}, CJRC \citep{duan2019cjrc}.
	\item {\bf Single Sentence Classification (SSC)}: ChnSentiCorp \citep{tan2008empirical}, THUCNews \citep{li2007scalable}.
	\item {\bf Sentence Pair Classification (SPC)}: XNLI \citep{conneau2018xnli}, LCQMC \citep{liu2018lcqmc}, BQ Corpus \citep{chen-etal-2018-bq}.
\end{itemize}

In order to make a fair comparison, for each dataset, we keep the same hyper-parameters (such as maximum length, warm-up steps, etc.) and only tune the initial learning rate from 1e-5 to 5e-5 for each task.
Note that the initial learning rates are tuned on original Chinese BERT, and it would be possible to achieve another gains by tuning the learning rate individually. 
We run the same experiment ten times to ensure the reliability of results.
The best initial learning rate is determined by selecting the best average development set performance.
We report the maximum and average scores to both evaluate the peak and average performance.

For all models except for ELECTRA, we use the same initial learning rate setting for each task, as depicted in Table \ref{hyper}.
For ELECTRA models, we use a universal initial learning rate of 1e-4 for base-level models and 5e-5 for large-level models as suggested in \citet{clark2020electra}.

As the pre-training data are quite different among various existing Chinese pre-trained language models, such as ERNIE \citep{sun2019ernie}, ERNIE 2.0 \citep{sun2019ernie2}, NEZHA \citep{wei2019nezha}, we only compare BERT \citep{devlin-etal-2019-bert}, BERT-wwm, BERT-wwm-ext, RoBERTa-wwm-ext, RoBERTa-wwm-ext-large, ELECTRA, along with our MacBERT to ensure relatively fair comparisons among different models, where all models are trained by ourselves except for the original Chinese BERT by \citet{devlin-etal-2019-bert}.
We carried out experiments under TensorFlow framework \citep{abadi2016tensorflow} with slight modifications to the fine-tuning scripts\footnote{https://github.com/google-research/bert} provided by \citet{devlin-etal-2019-bert} to better adapt to Chinese.

\section{Results}
\subsection{Machine Reading Comprehension}
Machine Reading Comprehension (MRC) is a representative document-level modeling task which requires to answer the questions based on the given passages.
We mainly test these models on three datasets: CMRC 2018, DRCD, and CJRC.
\begin{itemize}
	\item {\bf CMRC 2018}: A span-extraction machine reading comprehension dataset, which is similar to SQuAD \citep{rajpurkar-etal-2016} that extract a passage span for the given question.
	\item {\bf DRCD}: This is also a span-extraction MRC dataset but in Traditional Chinese. 
	\item {\bf CJRC}: Similar to CoQA \citep{reddy2019coqa}, which has yes/no questions, no-answer questions, and span-extraction questions. The data is collected from Chinese law judgment documents. Note that we only use {\tt small-train-data.json} for training. 
 \end{itemize}

\begin{table}[t]
\tiny
\begin{center}
\begin{tabular}{l c c c c}
\toprule
\multirow{2}*{\bf DRCD} & \multicolumn{2}{c}{\centering \bf Dev} & \multicolumn{2}{c}{\centering \bf Test} \\
& \bf EM & \bf F1 & \bf EM & \bf F1 \\
\midrule
BERT   				& 83.1 \tiny(82.7) & 89.9 \tiny(89.6) & 82.2 \tiny(81.6) & 89.2 \tiny(88.8) \\
BERT-wwm    		& 84.3 \tiny(83.4) & 90.5 \tiny(90.2) & 82.8 \tiny(81.8) & 89.7 \tiny(89.0) \\
BERT-wwm-ext		& 85.0 \tiny(84.5) & 91.2 \tiny(90.9) & 83.6 \tiny(83.0) & 90.4 \tiny(89.9)  \\
RoBERTa-wwm-ext 	& 86.6 \tiny(85.9) & 92.5 \tiny(92.2) & 85.6 \tiny(85.2) & 92.0 \tiny(91.7)  \\
ELECTRA-base 		& 87.5 \tiny(87.0) & 92.5 \tiny(92.3) & 86.9 \tiny(86.6) & 91.8 \tiny(91.7) \\
\bf MacBERT-base & \bf 89.4 \tiny(89.2) & \bf 94.3 \tiny(94.1) & \bf 89.5 \tiny(88.7) & \bf 93.8 \tiny(93.5) \\
\midrule
ELECTRA-large 		& 88.8 \tiny(88.7) & 93.3 \tiny(93.2) & 88.8 \tiny(88.2) & 93.6 \tiny(93.2) \\
RoBERTa-wwm-ext-L & 89.6 \tiny(89.1) & 94.8 \tiny(94.4) & 89.6 \tiny(88.9) & 94.5 \tiny(94.1) \\
\bf MacBERT-large 			  & \bf 91.2 \tiny(90.8) & \bf 95.6 \tiny(95.3) & \bf 91.7 \tiny(90.9) & \bf 95.6 \tiny(95.3) \\
\bottomrule
\end{tabular}
\end{center}
\caption{\label{result-drcd} Results on DRCD (Traditional Chinese). }
\end{table}

\begin{table}[t]
\tiny
\begin{center}
\begin{tabular}{l c c c c }
\toprule
\multirow{2}*{\bf CJRC} & \multicolumn{2}{c}{\centering \bf Dev} & \multicolumn{2}{c}{\centering \bf Test} \\
& \bf EM & \bf F1 & \bf EM & \bf F1 \\
\midrule
BERT   			& 54.6 \tiny(54.0) & 75.4 \tiny(74.5) & 55.1 \tiny(54.1) & 75.2 \tiny(74.3) \\
BERT-wwm    	& 54.7 \tiny(54.0) & 75.2 \tiny(74.8) & 55.1 \tiny(54.1) & 75.4 \tiny(74.4) \\
BERT-wwm-ext	& 55.6 \tiny(54.8) & 76.0 \tiny(75.3) & 55.6 \tiny(54.9) & 75.8 \tiny(75.0) \\
RoBERTa-wwm-ext & 58.7 \tiny(57.6) & 79.1 \tiny(78.3) & 59.0 \tiny(57.8) & 79.0 \tiny(78.0) \\
ELECTRA-base & 59.0 \tiny(58.1) & 79.4 \tiny(78.5) & 59.3 \tiny(58.2) & 79.4 \tiny(78.3) \\
\bf MacBERT-base & \bf 60.4 \tiny(59.5) & \bf 80.3 \tiny(79.2) & \bf 60.3 \tiny(59.3) & \bf 79.8 \tiny(79.0)  \\
\midrule
ELECTRA-large & 61.9 \tiny(60.8) & 82.1 \tiny(81.2) & 62.3 \tiny(61.2) & 82.0 \tiny(80.7) \\ 
RoBERTa-wwm-ext-L & 62.1 \tiny(61.1) & \bf 82.4 \tiny(81.6) & 62.4 \tiny(61.4) & 82.2 \tiny(81.0) \\
\bf MacBERT-large & \bf 62.4 \tiny(61.3) & 82.3 \tiny(81.4) & \bf 62.9 \tiny(61.6) & \bf 82.5 \tiny(81.1) \\
\bottomrule
\end{tabular}
\end{center}
\caption{\label{result-cjrc} Results on CJRC. }
\end{table}

\begin{table*}[htbp]
\small
\begin{center}
\begin{tabular}{p{3.5cm} c c | c c}
\toprule
\bf Single Sentence & \multicolumn{2}{c}{\centering \bf ChnSentiCorp} & \multicolumn{2}{c}{\centering \bf THUCNews} \\
\bf Classification & \bf Dev & \bf Test & \bf Dev & \bf Test \\
\midrule
BERT     			& 94.7 \tiny(94.3) & 95.0 \tiny(94.7) & 97.7 \tiny(97.4) & \bf 97.8 \tiny(97.6) \\
BERT-wwm     	& 95.1 \tiny(94.5) & 95.4 \bf\tiny(95.0) & 98.0 \tiny(97.6) & \bf 97.8 \tiny(97.6) \\
BERT-wwm-ext		& {\bf 95.4} \tiny(94.6) & 95.3 \tiny(94.8) & 97.7 \tiny(97.5) & 97.7 \tiny(97.5) \\
RoBERTa-wwm-ext 	& 94.9 \tiny(94.6) & {\bf 95.6} \tiny(94.9) & {\bf 98.3} \tiny(97.9) & 97.8 \tiny(97.5) \\
ELECTRA-base 	& 93.8 \tiny(93.0) & 94.5 \tiny(93.5) & 98.1 \tiny(97.9) & 97.8 \tiny(97.5) \\  
\bf MacBERT-base 	& 95.2 \bf\tiny(94.8) & {\bf 95.6} \tiny(94.9) & 98.2 \bf \tiny(98.0) &  97.7 \tiny(97.5)  \\	
\midrule
ELECTRA-large 	& 95.2 \tiny(94.6) & 95.3 \tiny(94.8) & 98.2 \bf\tiny(97.8) & 97.8 \tiny(97.6) \\
RoBERTa-wwm-ext-large & {\bf 95.8} \tiny(94.9) & 95.8 \tiny(94.9) & {\bf 98.3} \tiny(97.7) & 97.8 \tiny(97.6) \\
\bf MacBERT-large 	& 95.7 \bf \tiny(95.0) & {\bf 95.9} \bf\tiny(95.1) & 98.1 \bf\tiny(97.8) & \bf 97.9 \tiny(97.7) \\
\bottomrule
\end{tabular}
\end{center}
\caption{\label{result-spm} Results on single sentence classification tasks: ChnSentiCorp and THUCNews. }
\end{table*}

\begin{table*}[htbp]
\small
\begin{center}
\begin{tabular}{p{3.5cm} c c | c c | c c}
\toprule
\bf Sentence Pair & \multicolumn{2}{c}{\centering \bf XNLI} & \multicolumn{2}{c}{\centering \bf LCQMC} & \multicolumn{2}{c}{\centering \bf BQ Corpus} \\
\bf Classification & \bf Dev & \bf Test & \bf Dev & \bf Test  & \bf Dev & \bf Test \\
\midrule
BERT     			& 77.8 \tiny(77.4) & 77.8 \tiny(77.5) & 89.4 \tiny(88.4) & 86.9 \tiny(86.4) 	& 86.0 \tiny(85.5) 	& 84.8 \tiny(84.6) \\
BERT-wwm     	& 79.0 \tiny(78.4) & 78.2 \tiny(78.0) & 89.4 \tiny(89.2) & 87.0 \tiny(86.8)  & 86.1 \bf\tiny(85.6) 	& 85.2 \bf\tiny(84.9) \\
BERT-wwm-ext	& 79.4 \tiny(78.6) & 78.7 \tiny(78.3)  & 89.6 \tiny(89.2) & 87.1 \tiny(86.6) & {\bf 86.4} \tiny(85.5)  & {\bf 85.3} \tiny(84.8) \\
RoBERTa-wwm-ext & 80.0 \tiny(79.2) & 78.8 \tiny(78.3)  & 89.0 \tiny(88.7) & 86.4 \tiny(86.1) & 86.0 \tiny(85.4) & 85.0 \tiny(84.6) \\
ELECTRA-base 	& 77.9 \tiny(77.0) & 78.4 \tiny(77.8) & \bf 90.2 \tiny(89.8) & \bf 87.6 \tiny(87.3) & 84.8 \tiny(84.7) & 84.5 \tiny(84.0) \\  
\bf MacBERT-base & \bf 80.3 \tiny(79.7) & \bf 79.3 \tiny(78.8) & 89.5 \tiny(89.3) & 87.0 \tiny(86.5) & 86.0 \tiny(85.5) & 85.2 \bf\tiny(84.9) \\
\midrule
ELECTRA-large 	& 81.5 \tiny(80.8) & 81.0 \bf\tiny(80.9) & \bf 90.7 \tiny(90.4) & 87.3 \bf\tiny(87.2) & \bf 86.7 \tiny(86.2) & 85.1 \tiny(84.8) \\
RoBERTa-wwm-ext-large & 82.1 \tiny(81.3) & 81.2 \tiny(80.6)  & 90.4 \tiny(90.0) & 87.0 \tiny(86.8) & 86.3 \tiny(85.7) & {\bf 85.8} \tiny(84.9) \\
\bf MacBERT-large & \bf 82.4 \tiny(81.8) & {\bf 81.3} \tiny(80.6) & 90.6 \tiny(90.3) & {\bf 87.6} \tiny(87.1) & 86.2 \bf \tiny(85.7) & 85.6 \bf \tiny(85.0) \\
\bottomrule
\end{tabular}
\end{center}
\caption{\label{result-spm} Results on sentence pair classification tasks: XNLI, LCQMC, and BQ Corpus. }
\end{table*}

The results are depicted in Table \ref{result-cmrc2018}, \ref{result-drcd}, \ref{result-cjrc}.
Using additional pre-training data will result in further improvement, as shown in the comparison between BERT-wwm and BERT-wwm-ext.
This is why we use extended data for RoBERTa, ELECTRA, and MacBERT.
Moreover, the proposed MacBERT yields significant improvements on all reading comprehension datasets.
It is worth mentioning that our MacBERT-large could achieve a state-of-the-art F1 of 60\% on the challenge set of CMRC 2018, which requires deeper text understanding.

Also, it should be noted that though DRCD is a traditional Chinese dataset, training with additional large-scale simplified Chinese could also have a great positive effect.
As simplified and traditional Chinese share many identical characters, using a powerful pre-trained language model with only a few traditional Chinese data could also bring improvements without converting traditional Chinese characters into simplified ones.

Regarding CJRC, where the text is written in professional ways regarding Chinese laws, BERT-wwm shows moderate improvement over BERT but not that salient, indicating that further domain adaptation is needed for the fine-tuning tasks on non-general domains.
However, by increasing general training data will result in improvement, suggesting that when there is no enough domain data, we could also use large-scale general data as a remedy.

\subsection{Single Sentence Classification}
For single sentence classification tasks, we select ChnSentiCorp and THUCNews datasets.
We use the ChnSentiCorp dataset for evaluating sentiment classification, where the text should be classified into either a positive or negative label.
THUCNews is a dataset that contains news in different genres, where the text is typically very long.
In this paper, we use a version that contains 50K news in 10 domains (evenly distributed), including sports, finance, technology, etc.\footnote{https://github.com/gaussic/text-classification-cnn-rnn}
The results show that our MacBERT could give moderate improvements over baselines, as these datasets have already reached very high accuracies.

\subsection{Sentence Pair Classification}
For sentence pair classification tasks, we use XNLI data (Chinese portion), Large-scale Chinese Question Matching Corpus (LCQMC), and BQ Corpus, which require to input two sequences and predict their relations.
We can see that MacBERT outperforms other models, but the improvements were moderate, with a slight improvement on the average score, but the peak performance is not as good as RoBERTa-wwm-ext-large.
We suspect that these tasks are less sensitive to the subtle difference of the input than the reading comprehension tasks. As sentence pair classification only needs to generate a unified representation of the whole input and thus results in a moderate improvement.

\section{Discussion}
While our models achieve significant improvements on various Chinese tasks, we wonder where the essential components of the improvements from.
To this end, we carried out detailed ablations on MacBERT to demonstrate their effectiveness, and we also compare the claims of the existing pre-trained language models in English to see if their modification still holds true in another language.

\begin{table*}[htbp]
\small
\begin{center}
\begin{tabular}{l cc | cc | cc | c c | c c c | c}
\toprule
& \multicolumn{2}{c}{\centering \bf CMRC 2018} & \multicolumn{2}{c}{\centering \bf DRCD}  & \multicolumn{2}{c}{\centering \bf CJRC} & {\bf CSC} & {\bf THUC} & {\bf XNLI} & {\bf LC} &  {\bf BQ} & \multirow{2}*{\bf AVG}  \\
& \bf EM & \bf F1 & \bf EM & \bf F1 & \bf EM & \bf F1  & \bf ACC & \bf ACC & \bf ACC & \bf ACC & \bf ACC \\
\midrule
MacBERT-large		& 74.8 & 90.7 & 91.7 & 95.6 & 62.9 & 82.5 & 95.9 & 97.9 & 81.3 & 87.6 & 85.6 & 87.18 \\
SOP $\rightarrow$ NSP	& 74.5 & 90.6 & 91.5 & 95.5 & 62.4 & 82.3 & 96.0 & 97.8 & 81.2 & 87.4 & 85.2 & 87.00 \\
w/o SOP 				& 74.4 & 90.6 & 91.0 & 95.4 & 62.2 & 82.1 & 95.8 & 97.8 & 81.1 & 87.4 & 85.2 & 86.89 \\
\midrule
w/o Mac	& 74.2 & 90.1 & 91.2 & 95.4 & 62.2 & 82.3 & 95.7 & 97.8 & 81.2 & 87.4 & 85.3 & 86.88  \\
w/o NM 	& 74.0 & 89.8 & 90.9 & 95.1 & 62.1 & 82.0 & 95.9 & 97.9 & 81.3 & 87.5 & 85.6 & 86.89 \\
RoBERTa-large 	& 74.2 & 90.6 & 89.6 & 94.5 & 62.4 & 82.2 & 95.8 & 97.8 & 81.2 & 87.0 & 85.8 & 86.79 \\
\bottomrule
\end{tabular}
\end{center}
\caption{\label{ablation} Ablations of MacBERT-large on different fine-tuning tasks.}
\end{table*}

\subsection{Effectiveness of MacBERT}\label{effect-macbert}
We carried out ablations to examine the contributions of each component in MacBERT, which was thoroughly evaluated in all fine-tuning tasks.
The results are shown in Table \ref{ablation}. 
The overall average scores are obtained by averaging the test scores of each task (EM and F1 metrics are averaged before the overall averaging).
From a general view, removing any component in MacBERT will result in a decline in the average performance, suggesting that all modifications contribute to the overall improvements.
Specifically, the most effective modifications are the N-gram masking and similar word replacement, which are the modifications on the masked language model task.
When we compare N-gram masking and similar word replacement, we could see clear pros and cons, where N-gram masking seems to be more effective in text classification tasks, and the performance of reading comprehension tasks seems to benefit more from the similar word replacement task.
By combining these two tasks will compensate each other and have a better performance on both genres.

The NSP task does not show as much importance as the MLM task, demonstrating that it is much more important to design a better MLM task to fully unleash the text modeling power.
Also, we compared the next sentence prediction \citep{devlin-etal-2019-bert} and sentence order prediction \citep{lan2019albert} task to better judge which one is much powerful.
The results show that the sentence order prediction task indeed shows better performance than the original NSP, though it is not that salient.
The SOP task requires identifying the correct order of the two sentences rather than using a random sentence, which is much easy for the machine to identify. 
Removing the SOP task will result in noticeable declines in reading comprehension tasks compared to the text classification tasks, which suggests that it is necessary to design an NSP-like task to learn the relations between two segments (for example, passage and question in reading comprehension task).

\begin{figure}[t]
\centering
\subfigure{
\begin{minipage}[htbp]{0.95\linewidth}
\centering
\includegraphics[width=1\columnwidth]{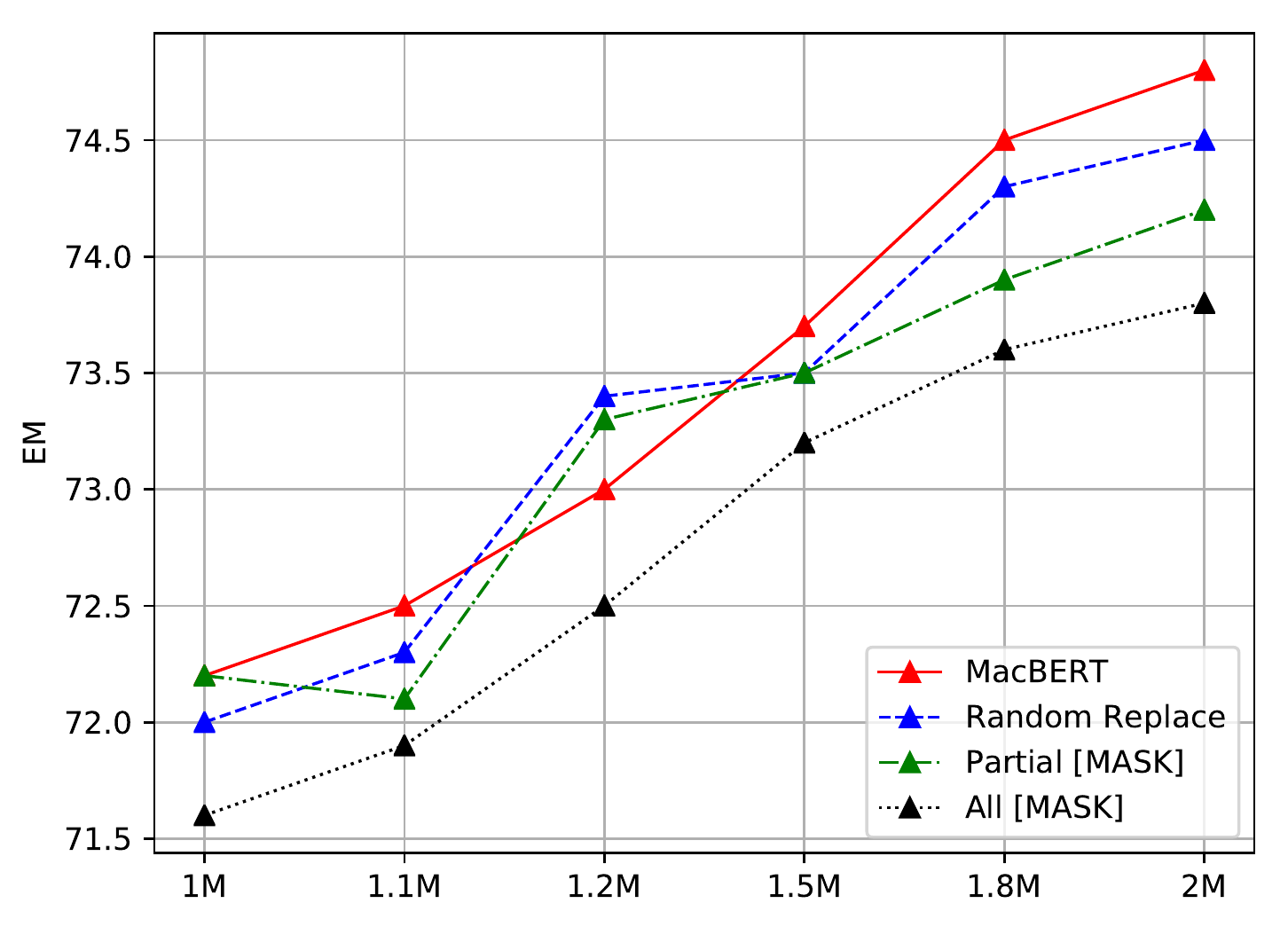} \\
\includegraphics[width=1\columnwidth]{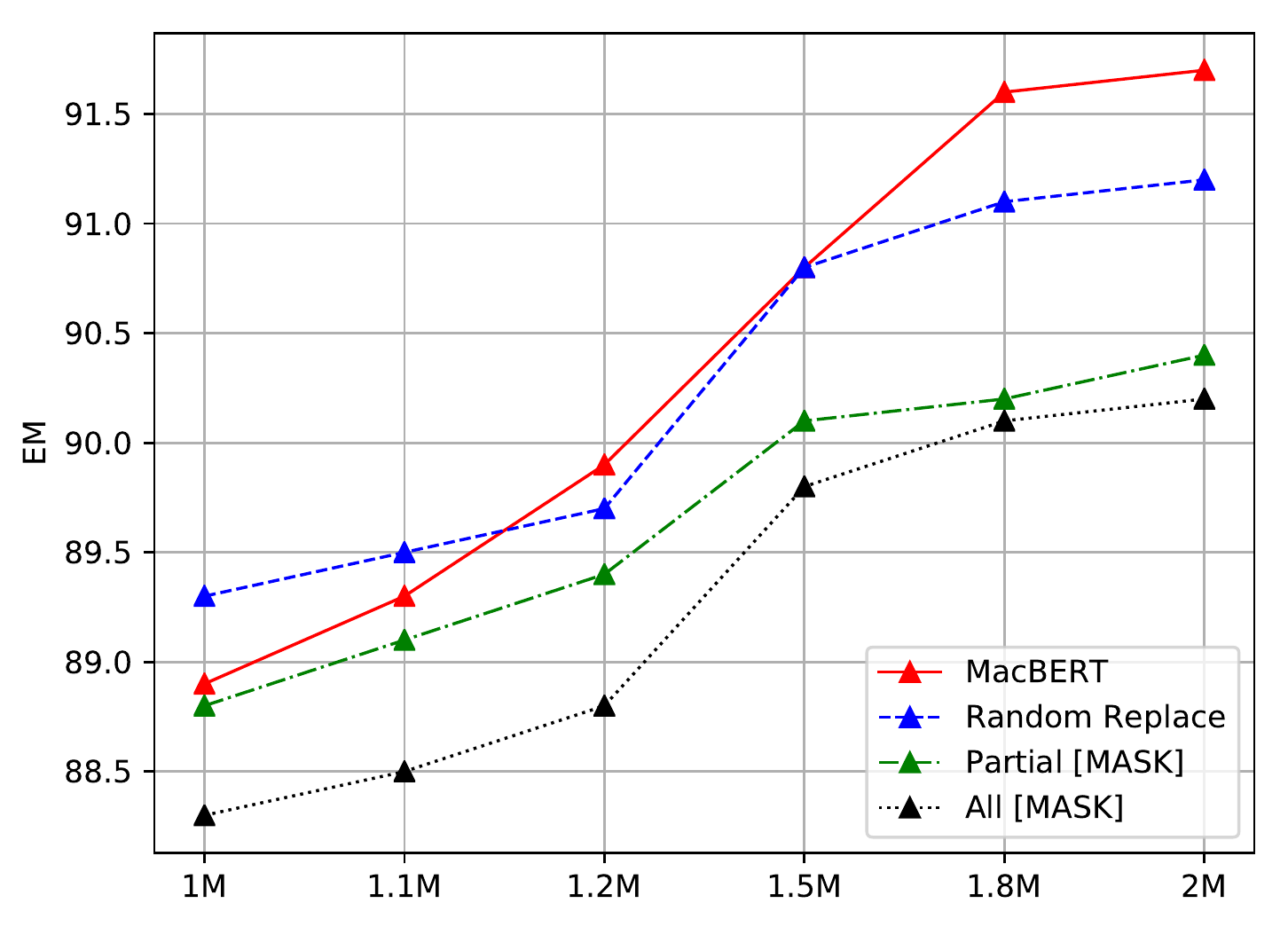}
\end{minipage}%
}%
\centering
  \caption{\label{discussion} Results of different MLM tasks on CMRC 2018 and DRCD.}
\end{figure}

\subsection{Investigation on MLM Task}
As said in the previous section, the dominant pre-training task is the masked language model and its variants.
The masked language model task relies on two sides: 1) the selection of the tokens to be masked, and 2) the replacement of the selected tokens.
In the previous section, we have demonstrated the effectiveness of the selection of the masking tokens, such as the whole word masking or N-gram masking, etc.
Now we are going to investigate how the replacement of the selected tokens will affect the performance of the pre-trained language models.
In order to investigate this problem, we plot the CMRC 2018 and DRCD performance at different pre-training steps.
Specifically, we follow the original masking percentage 15\% of the input sequence, in which 10\% masked tokens remain the same.
In terms of the remaining 90\% masked tokens, we classify into four categories.
\begin{itemize}
	\item {\bf MacBERT}: 80\% tokens replaced into their similar words, and 10\% replaced into random words. 
	\item {\bf Random Replace}: 90\% tokens replaced into random words.
	\item {\bf Partial Mask}: original BERT implementation, with 80\% tokens replaced into {\tt [MASK]} tokens, and 10\% replaced into random words.	
	\item {\bf All Mask}: 90\% tokens replaced with {\tt [MASK]} tokens.
\end{itemize}
  
We only plot the steps from 1M to 2M to show stabler results than the first 1M steps.
The results are depicted in Figure \ref{discussion}.
The pre-training models that rely on mostly using {\tt [MASK]} for masking purpose (i.e., partial mask and all mask) results in worse performances, indicating that the discrepancy of the pre-training and fine-tuning is an actual problem that affects the overall performance.
Among which, we also noticed that if we do not leave 10\% as original tokens (i.e., identity projection), there is also a consistent decline, indicating that masking with {\tt [MASK]} token is less robust and vulnerable to the absence of identity projection for negative sample training.

To our surprise, a quick fix, that is to abandon the {\tt [MASK]} token completely and replace all 90\% masked tokens into random words, yields consistent improvements over {\tt [MASK]}-dependent masking strategies. 
This also strengthens the claims that the original masking method that relies on the {\tt [MASK]} token, which never appears in the fine-tuning task, will result in a discrepancy and worse performance.
To make this more delicate, in this paper, we propose to use similar words for masking purpose, instead of randomly pick a word from the vocabulary, as random word will not fit in the context and may break the naturalness of the language model learning, as traditional N-gram language model is based on natural sentence rather than a manipulated influent sentence.
However, if we use similar words for masking purposes, the fluency of the sentence is much better than using random words, and the whole task transforms into a grammar correction task, which is much more natural and without the discrepancy of the pre-training and fine-tuning stage.
From the chart, we can see that the MacBERT yields the best performance among the four variants, which verifies our assumptions.

\section{Conclusion}
In this paper, we revisit pre-trained language models in Chinese to see if the techniques in these state-of-the-art models generalize well in a different language other than English only.
We created Chinese pre-trained language model series and proposed a new model called MacBERT, which modifies the masked language model (MLM) task as a language correction manner and mitigates the discrepancy of the pre-training and fine-tuning stage.
Extensive experiments are conducted on various Chinese NLP datasets, and the results show that the proposed MacBERT could give significant gains in most of the tasks, and detailed ablations show that more focus should be made on the MLM task rather than the NSP task and its variants, as we found that NSP-like task does not show a landslide advantage over one another.
With the release of the Chinese pre-trained language model series, we hope it will further accelerate the natural language processing in the Chinese research community.

In the future, we would like to investigate an effective way to determine the masking ratios instead of heuristic ones to further improve the performance of the pre-trained language models.

\section*{Acknowledgments}\label{ack}
We would like to thank all anonymous reviewers and senior program members for their thorough reviewing and providing constructive comments to improve our paper. 
The first author was partially supported by the Google TensorFlow Research Cloud (TFRC) program for Cloud TPU access.
This work was supported by the National Natural Science Foundation of China (NSFC) via grant 61976072, 61632011, and 61772153.

\bibliography{emnlp2020}
\bibliographystyle{acl_natbib}

\appendix
\section{Appendix}
\subsection{XLNet Results on Machine Reading Comprehension Tasks}
Following official XLNet implementation, we trained a sentencepiece vocabulary of 32,000 and used it for word segmentation.
We use exactly the same pre-training data as those marked as `ext' models with 5.4B training tokens.
We mainly implemented XLNet-base (12-layers, 768 hidden dimension) and XLNet-mid (24-layers, 768 hidden dimension).
The pre-training of XLNet-mid and XLNet-base was done on a single Cloud TPU v3 for 2M/4M steps with a batch size of 32, respectively.
\begin{table*}[t]
\begin{center}
\begin{tabular}{l c c c c c c}
\toprule
\multirow{2}*{\bf CMRC 2018} & \multicolumn{2}{c}{\centering \bf Dev} & \multicolumn{2}{c}{\centering \bf Test} & \multicolumn{2}{c}{\centering \bf Challenge} \\
& \bf EM & \bf F1 & \bf EM & \bf F1  & \bf EM & \bf F1  \\
\midrule
BERT   			& 65.5 \tiny(64.4) & 84.5 \tiny(84.0) & 70.0 \tiny(68.7) & 87.0 \tiny(86.3)  & 18.6 \tiny(17.0) & 43.3 \tiny(41.3) \\
\bf XLNet-base     	& 65.2 \tiny(63.0) & 86.9 \tiny(85.9) & 67.0 \tiny(65.8) & 87.2 \tiny(86.8)  & 25.0 \tiny(22.7) & 51.3 \tiny(49.5) \\
\bf XLNet-mid		& 66.8 \tiny(66.3) & 88.4 \tiny(88.1) & 69.3 \tiny(68.5) & 89.2 \tiny(88.8) & 29.1 \tiny(27.1) & 55.8 \tiny(54.9) \\
\bottomrule
\end{tabular}
\end{center}
\caption{\label{result-xlnet-cmrc2018} Results of XLNet on CMRC 2018 (Simplified Chinese). }
\end{table*}

\begin{table*}[t]
\begin{center}
\begin{tabular}{l c c c c }
\toprule
\multirow{2}*{\bf DRCD} & \multicolumn{2}{c}{\centering \bf Dev} & \multicolumn{2}{c}{\centering \bf Test} \\
& \bf EM & \bf F1 & \bf EM & \bf F1 \\
\midrule
BERT   				& 83.1 \tiny(82.7) & 89.9 \tiny(89.6) & 82.2 \tiny(81.6) & 89.2 \tiny(88.8) \\
\bf XLNet-base    		& 83.8 \tiny(83.2) & 92.3 \tiny(92.0) & 83.5 \tiny(82.8) & 92.2 \tiny(91.8) \\
\bf XLNet-mid			& 85.0 \tiny(84.5) & 91.2 \tiny(90.9) & 85.5 \tiny(84.8) & 93.6 \tiny(93.2)  \\
\bottomrule
\end{tabular}
\end{center}
\caption{\label{result-xlnet-drcd} Results of XLNet on DRCD (Traditional Chinese). }
\end{table*}

The results on CMRC 2018 and DRCD are shown in Table \ref{result-xlnet-cmrc2018} and \ref{result-xlnet-drcd}. 
The results show that these XLNet models could achieve moderate improvements over BERT, and the improvements are not consistent on each subset.

We also carried out experiments on text classification task, such as XNLI, but the XLNet-mid could only gives near 74\% on the test set, while the BERT-base could reach an accuracy of 77.8\%. 
We haven't figured out the exact issues and also did not find other successful Chinese XLNet in the community.
We will investigate the issue and will update these results once we figure it out through our open-source implementation repository.

\end{CJK*}
\end{document}